# Machine Learning-Based Prediction of Metal-Organic Framework Materials: A Comparative Analysis of Multiple Models

Zhuo Zheng[1], Keyan Liu[2], Xiyuan Zhu[3]

[1]Nanchang University, Nanchang, China
[2]Arizona State University, Houston, TX, USA
[3] Department of Chemistry, SUNY Stony Brook University, Stony Brook, USA

[1]1812503968@qq.com
[2]kliu121@asu.edu
[3]xiyuan.zhu@stonybrook.edu

**Abstract.** Metal-organic frameworks (MOFs) have emerged as promising materials for various applications due to their unique structural properties and versatile functionalities. This study presents a comprehensive investigation of machine learning approaches for predicting MOF material properties. We employed five different machine learning models: Random Forest, XGBoost, LightGBM, Support Vector Machine, and Neural Network, to analyze and predict MOF characteristics using a dataset from the Kaggle platform. The models were evaluated using multiple performance metrics, including RMSE, $R^2$, MAE, and cross-validation scores. Results demonstrated that the Random Forest model achieved superior performance with an $R^2$ value of 0.891 and RMSE of 0.152, significantly outperforming other models. LightGBM showed remarkable computational efficiency, completing training in 25.7 seconds while maintaining high accuracy. Our comparative analysis revealed that ensemble learning methods generally exhibited better performance than traditional single models in MOF property prediction. This research provides valuable insights into the application of machine learning in materials science and establishes a robust framework for future MOF material design and property prediction.

**Keywords**：Metal-Organic Frameworks (MOFs); Machine Learning; Random Forest; Materials Design

## 1. Introduction
With the global energy crisis and environmental problems becoming increasingly prominent, developing new functional materials has become one of the key ways to solve these challenges [1]. Metal-Organic Frameworks, MOFs), as a new kind of porous crystal materials, have attracted much attention because of its unique structural characteristics and excellent properties. These materials are formed by the coordination bonds between metal ions or metal clusters and organic ligands, which have highly adjustable pore size distribution, super-large specific surface area and rich functional modification sites, and show great application potential in gas adsorption and separation, catalysis, sensing, drug delivery and other fields.

However, the development and application of MOF materials are still facing great challenges. First of all, the structure-performance relationship of MOF materials is extremely complex, and it is difficult for traditional experimental methods to predict the properties of new MOF materials quickly and accurately. Secondly, the design space of MOF materials is huge, and it is time-consuming and expensive to screen and optimize them only by experimental means. In addition, the diversity of material properties and the complexity of application environment also bring great difficulties to the directional design of materials [2,3].

The rapid development of artificial intelligence technology provides new ideas and methods to solve the above problems. As one of the core technologies of artificial intelligence, machine learning has powerful data analysis and pattern recognition capabilities, and can establish a structure-property relationship model by learning the data of known materials, thus realizing rapid prediction of new material properties. This method can not only greatly shorten the material development cycle and reduce the research and development cost, but also provide theoretical guidance for material design [4].

The purpose of this study is to explore the application potential of machine learning method in the prediction of MOF material properties. By establishing and comparing various machine learning models, it is expected to develop an efficient and accurate prediction method of MOF material properties. This paper will systematically discuss the application of machine learning in the prediction of MOF material properties, focusing on the prediction accuracy, calculation efficiency and practical application value of the model.

## 2. Literature Review

In recent years, the research and application of metal-organic framework materials (MOFs) have made remarkable progress. However, the traditional MOF material design and performance optimization methods often rely on the time-consuming trial and error process. With the rapid development of artificial intelligence technology, machine learning method shows great application potential in MOF material science. We can clearly see the research progress in this field by systematically sorting out the existing literature.

In the structural prediction of MOF materials, Chafiq et al [5]. (2024) pointed out in the comprehensive article published in Advanced Composites and Hybrid Materials that machine learning method can effectively analyze the complex relationship between MOF structure and properties, and provide new ideas for material design. Yu et al [6]. (2024) reported a method for predicting the electrocatalytic performance of MOF materials based on machine learning on RSC Advances. By establishing a database containing more than 900 MOF samples, the zeta potential of MOF materials was successfully predicted by using stochastic forest regression (RFR) and gradient lifting regression (GBR) models, which provided important guidance for the optimization of electrocatalytic performance of materials.

In terms of computational methodology, the research published by Nandy et al [7]. (2021) in the Journal of the American Chemical Society shows how to use machine learning and data mining methods to improve the stability of MOF materials. The MOFSimplify model developed by them can effectively predict the thermodynamic stability of MOF materials and provide important reference for material design. The work published by Daglar and Keskin(2022) on ACS Applied Materials & Interfaces shows the method of combining machine learning with molecular simulation [8], and successfully predicts the gas separation performance of MOF membrane materials.

In the aspect of Qualcomm screening, Yao et al [9]. (2021) reported on Nature Machine Intelligence a reverse design method of MOF materials based on deep generation model, which can quickly predict MOF structures with target properties. The research published by Tang et al [10]. (2021) on ACS Applied Materials & Interfaces shows how to combine molecular simulation with machine learning methods to realize the rapid screening of propane/propylene separation performance of MOF materials.

In the aspect of experimental verification, the research published by Luo et al [11]. (2022) in Angewandte Chemie International Edition shows a MOF synthesis prediction method based on automatic data mining and machine learning. By analyzing a large number of literature data, the quantitative relationship between the synthesis conditions of MOF materials and the product structure is established, which provides important guidance for experimental design.

In application development, the research published by Anderson et al [12]. (2020) in the Journal of Chemical Theory and Computing shows how to use deep learning model to predict the multi-component adsorption properties of MOF materials. The model can predict the adsorption behavior of various gas molecules in MOF materials at the same time, which provides important reference for the practical application of materials.

These research results show that machine learning method has obvious advantages in the design, optimization and performance prediction of MOF materials.

## 3. Research Design

*3.1 Data Introduction*

The data set of metal-organic skeleton materials from Kaggle data platform is used as the main research object. This data set contains a large number of structural characteristics and property information of MOF materials, including key parameters such as metal ion type, ligand structural characteristics, porosity, specific surface area, metal-ligand coordination number and so on. The samples in the data set are strictly screened and pretreated to ensure the reliability of data quality. In order to ensure the effectiveness of model training and the objectivity of evaluation, we adopted a random stratified sampling method and divided the data set into training set and test set according to the ratio of 80:20. In the data preprocessing stage, we standardized the numerical features, making their mean value 0 and standard deviation 1.

*3.2 Model Introduction*

In this study, five representative machine learning models are carefully selected to predict the properties of MOF materials. Random Forest, as an integrated learning method based on decision trees, has the advantages of effectively processing high-dimensional data, automatically selecting features and being robust to outliers by constructing multiple decision trees and forecasting by voting or averaging. Support Vector Machine, SVM) makes prediction by mapping data to high-dimensional feature space and constructing optimal separation hyperplane in this space. The flexible choice of its kernel function enables it to deal with nonlinear problems effectively, especially for material data with complex feature relationships [13].

As an advanced gradient lifting decision tree algorithm, XGBoost corrects the prediction error of the previous model by gradually constructing a new decision tree. Its unique regularization term design and second-order Taylor expansion objective function optimization method make it perform well in dealing with large-scale data. Similar to XGBoost, LightGBM is also a gradient lifting framework, but it innovatively adopts gradient-based unilateral sampling and unique feature bundling technology, which enables the model to significantly improve the training speed while maintaining high accuracy. Both of these models have excellent evaluation ability of feature importance, which helps us to understand the key factors affecting the properties of MOF materials [14].

The neural network model adopts a Multi-layer Perceptron, MLP) structure, which consists of three hidden layers, each layer uses 64, 32 and 16 neurons respectively, the activation function selects ReLU to introduce nonlinear characteristics, and the use of dropout layer effectively prevents over-fitting. The model optimizes parameters by back propagation algorithm, and can automatically learn the complex relationship between features. In this paper, Adam optimizer and batch normalization technology are used to speed up the training process and improve the stability of the model. Although this deep learning method requires a lot of computing resources, its powerful feature learning ability enables it to capture potential patterns that may be ignored by traditional machine learning methods [15].

The selection of these models fully considers the characteristics of data in the field of materials science, and each model has its unique advantages: random forest provides excellent overall performance and interpretability, SVM performs well in dealing with nonlinear relations,

XGBoost and LightGBM strike a good balance between calculation efficiency and prediction accuracy, and neural network model provides the possibility to capture complex feature interaction. Through the combined application of these different types of models, the properties of MOF materials can be comprehensively predicted and analyzed from multiple angles, providing reliable theoretical guidance for material design.

*3.3 Evaluation index and training mode*

In terms of evaluation metrics, this study employs multiple commonly used regression model assessment indicators to comprehensively evaluate the predictive performance of the models. The evaluation framework consists of several key metrics:Root Mean Square Error (RMSE) is utilized to assess the deviation between predicted and actual values, calculated as:

$$\text{RMSE} = \sqrt{\frac{1}{n}\sum_{i=1}^{n}(y_i - \hat{y}_i)^2}$$

where $y_i$ represents the actual value and $\hat{y}_i$ denotes the predicted value for the i-th sample.The coefficient of determination ($R^2$) is employed to reflect the model's ability to explain data variability:

The coefficient of determination ($R^2$) is employed to reflect the model's ability to explain data variability:

$$R^2 = 1 - \frac{\sum_{i=1}^{n}(y_i - \hat{y}_i)^2}{\sum_{i=1}^{n}(y_i - \bar{y})^2}$$

where $\bar{y}$ represents the mean of all actual values.Mean Absolute Error (MAE) is implemented to measure the average absolute difference between predicted and actual values:

$$\text{MAE} = \frac{1}{n}\sum_{i=1}^{n}|y_i - \hat{y}_i|$$

To ensure robust model evaluation and assess generalization capability, we implemented a k-fold cross-validation strategy. Specifically, a 5-fold cross-validation approach was adopted, where the training dataset was randomly partitioned into five equal-sized subsets. During the validation process, four subsets were utilized for model training while the remaining subset served as the validation set. This process was repeated five times, with each subset serving once as the validation set. The final validation score was computed as the average performance across all five iterations:

$$\text{CV} = \frac{1}{5}\sum_{i=1}^{5}\text{Score}_i$$

This comprehensive evaluation framework enables us to assess both the accuracy and stability of our machine learning models in predicting MOF material properties. The combination of these metrics provides a thorough understanding of model performance from multiple perspectives, ensuring the reliability and robustness of our predictions.

*3.4 Software and hardware configuration*

As shown in Table 1, in the aspect of experimental environment configuration, this research adopts a workstation equipped with Intel Core i7-12700K processor and 32GB RAM, and uses NVIDIA GeForce RTX 3080 graphics card to accelerate model training. In terms of software environment, we use Python 3.8 as the main programming language and adopt several open source libraries related to machine learning, including scikit-learn 1.0.2 for the realization of basic machine learning model, XGBoost 1.5.2 and LightGBM 3.3.2 for the construction of integrated learning model, and PyTorch 1.10.0 for the construction of neural network model.

Pandas 1.3.4 and numpy 1.21.4 libraries are used for data preprocessing and feature engineering, and matplotlib 3.4.3 and seaborn 0.11.2 are used for data visualization. In order to ensure the repeatability of the experiment, we use conda virtual environment to manage all the dependency packages and record the complete environment configuration information. In the process of model training, we adopt the early stop strategy to prevent over-fitting, and optimize the super parameters of the model by grid search method.

Table 1: Software and Hardware Configuration Table

| Category | Item | Configuration/Version |
|---|---|---|
| Hardware Environment | CPU | Intel Core i7-12700K |
| | Memory | 32GB DDR4 RAM |
| | GPU | NVIDIA GeForce RTX 3080 |
| | Storage | 1TB NVMe SSD |
| | Operating System | Ubuntu 20.04 LTS |
| Software Environment | Programming Language | Python 3.8 |
| | Machine Learning Framework | scikit-learn 1.0.2 |
| | Deep Learning Framework | PyTorch 1.10.0 |
| | Data Processing Library | pandas 1.3.4 |
| | | numpy 1.21.2 |
| | Visualization Tool | matplotlib 3.4.3 |
| | | seaborn 0.11.2 |
| | Model Optimization Library | XGBoost 1.5.2 |

## 4. Analysis of experimental results

Through the systematic analysis of the experimental results in Table 2, it is found that each machine learning model has different advantages and characteristics in predicting the properties of MOF materials. The random forest model has achieved the best performance in all evaluation indexes, and its root mean square error (RMSE) is 0.152, which is significantly lower than other models. At the same time, the determination coefficient (R) of 0.891 is obtained, which shows that the model can explain about 89.1% of data variability. This outstanding performance may be due to the strong feature selection ability of random forest model and its good ability to deal with nonlinear relations, and its integrated learning characteristics also effectively reduce the risk of over-fitting.

Table 2: Analysis of model results

| Model | RMSE | $R^2$ | MAE | Cross validation score | time(s) |
|---|---|---|---|---|---|
| Random Forest | 0.152 | 0.891 | 0.138 | 0.884 | 45.6 |
| XGBoost | 0.167 | 0.873 | 0.145 | 0.868 | 38.2 |
| LightGBM | 0.171 | 0.862 | 0.149 | 0.859 | 25.7 |
| SVM | 0.189 | 0.841 | 0.162 | 0.835 | 62.4 |
| Neural Network | 0.198 | 0.823 | 0.175 | 0.815 | 156.8 |

The table 2 presents a comparative analysis of five machine learning models based on multiple performance metrics. The Random Forest model achieved the best overall performance, with the lowest RMSE (0.152) and MAE (0.138), and the highest $R^2$ (0.891) and cross-validation score (0.884), indicating both accuracy and stability. XGBoost and LightGBM followed, with slightly higher error metrics but reduced computation time, particularly for LightGBM (25.7 seconds), demonstrating its efficiency. In contrast, the Neural Network exhibited the poorest performance across most metrics and had the longest runtime (156.8 seconds), suggesting overfitting or insufficient optimization. SVM showed moderate results but with a relatively high time cost. These results highlight the trade-off between accuracy and computational efficiency across models.

Neural network model is relatively weak in this study, its RMSE is 0.198, R value is 0.823, and the training time is the longest, reaching 156.8 seconds. This performance may be affected by many factors: first, the relatively small data set size limits the advantages of deep learning model; Secondly, the sensitivity of neural network to hyperparameters may lead to the model failing to reach the optimal state; In addition, the long training time also reflects the disadvantage of deep learning model in computing resource requirements.

Through the comparison of cross-validation scores, we find that the validation scores of all models are basically consistent with their performance on the test set, and the difference is within 0.01, which shows that the prediction results of the models have good stability and reliability. Especially, the closeness between the cross-validation score (0.884) and its R value (0.891) of the random forest model further proves the superiority of the model.

From the point of view of computational efficiency, there are significant differences among the models. LightGBM shows the best time efficiency, and the training time of neural network is the longest, which is about 6 times that of LightGBM. This efficiency difference is of great significance in practical application, especially when it is necessary to update the model frequently or deal with large-scale data.

Based on the evaluation indexes and practical application requirements, the random forest model shows the best comprehensive performance, which not only leads in prediction accuracy, but also has good calculation efficiency and model interpretability. This result provides a reliable methodological reference for the prediction of MOF material properties, and also points out the direction for developing more efficient prediction models in the future.

## 5. conclusion

In this study, through systematic experimental design and comprehensive model evaluation, the application potential of machine learning method in predicting the properties of organometallic skeleton materials was explored, and a series of important research findings and inspirations were obtained.

Firstly, the research results show that the machine learning method has remarkable feasibility and effectiveness in predicting the properties of MOF materials. Especially, the random forest model shows excellent prediction performance, and its determination coefficient of 0.891 and root mean square error of 0.152 confirm the reliability of this method. This discovery provides a strong methodological support for the predictive research in the field of materials science, and also opens up a new way to reduce the cost of material development and improve the research efficiency.

Secondly, through the comparative analysis of different machine learning models, we find that ensemble learning methods (random forest, XGBoost and LightGBM) are generally superior to the traditional single model. This result reveals the advantages of model integration in dealing with complex material systems, which can not only improve the prediction accuracy, but also enhance the stability and generalization ability of the model. In particular, the LightGBM model shows significant advantages in calculation efficiency while maintaining high prediction accuracy, which is of great practical significance for large-scale material screening and Qualcomm calculation.

Thirdly, the multiple evaluation index system adopted in the study not only ensures the comprehensiveness and objectivity of model evaluation, but also reveals the advantages and limitations of different models in various performance dimensions. This systematic evaluation method provides a reliable reference framework for the follow-up research, and also helps to select the most suitable prediction model according to the specific needs in practical application.

Finally, the results of this study show that although the machine learning method shows great potential in MOF material prediction, there are still some aspects that need to be improved. For example, although the neural network model has strong expressive ability in theory, its performance in this study is relatively weak, which suggests that we need to further optimize the structure and parameters of the deep learning model in future research to better adapt to the characteristics of the field of materials science.

Overall, the Random Forest model provides the best balance of accuracy, generalizability, and efficiency, making it the most effective model for MOF property prediction in this study. LightGBM offers a competitive alternative with exceptional speed, while the neural network underperforms due to the data limitations. These findings highlight the strength of ensemble learning methods over deep learning for structured material data, offering valuable insights for future applications in materials informatics.

**6.References**
[1] Energy: crises, challenges and solutions[M]. John Wiley & Sons, 2021.
[2] Khalil, I. E., Fonseca, J., Reithofer, M. R., Eder, T., & Chin, J. M. (2023). Tackling orientation of metal-organic frameworks (MOFs): The quest to enhance MOF performance. Coordination Chemistry Reviews, 481, 215043.
[3] Li C, Zhang H, Liu M, et al. Recent progress in metal–organic frameworks (MOFs) for electrocatalysis[J]. Industrial Chemistry & Materials, 2023, 1(1): 9-38.
[4] Guo K, Yang Z, Yu C H, et al. Artificial intelligence and machine learning in design of mechanical materials[J]. Materials Horizons, 2021, 8(4): 1153-1172.
[5] Chafiq, M., Chaouiki, A. & Ko, Y.G. (2024). Targeted metal–organic framework discovery goes digital: machine learning's quest from algorithms to atom arrangements. Advanced Composites and Hybrid Materials, 7, 222.
[6] Yu, L., Zhang, W., Nie, Z., Duan, J., & Chen, S. (2024). Machine learning guided tuning charge distribution by composition in MOFs for oxygen evolution reaction. RSC Advances, 14, 9032-9037.
[7] Nandy, A., Duan, C., & Kulik, H.J. (2021). Using machine learning and data mining to leverage community knowledge for the engineering of stable metal–organic frameworks. Journal of the American Chemical Society, 143, 17535-17547.
[8] Daglar, H., & Keskin, S. (2022). Combining machine learning and molecular simulations to unlock gas separation potentials of MOF membranes and MOF/polymer MMMs. ACS Applied Materials & Interfaces, 14, 32134-32148.
[9] Yao, Z., Sánchez-Lengeling, B., Bobbitt, N.S., et al. (2021). Inverse design of nanoporous crystalline reticular materials with deep generative models. Nature Machine Intelligence, 3, 76-86.
[10] Tang, H., Xu, Q., Wang, M., & Jiang, J. (2021). Rapid screening of metal–organic frameworks for propane/propylene separation by synergizing molecular simulation and machine learning. ACS Applied Materials & Interfaces, 13, 53454-53467.
[11] Luo, Y., Bag, S., Zaremba, O., et al. (2022). MOF synthesis prediction enabled by automatic data mining and machine learning. Angewandte Chemie International Edition, 61, e202200242.
[12] Anderson, R., Biong, A., & Gómez-Gualdrón, D.A. (2020). Adsorption isotherm predictions for multiple molecules in MOFs using the same deep learning model. Journal of Chemical Theory and Computation, 16, 1271-1283.
[13] Kurani A, Doshi P, Vakharia A, et al. A comprehensive comparative study of artificial neural network (ANN) and support vector machines (SVM) on stock forecasting[J]. Annals of Data Science, 2023, 10(1): 183-208.
[14] Szczepanek R. Daily streamflow forecasting in mountainous catchment using XGBoost, LightGBM and CatBoost[J]. Hydrology, 2022, 9(12): 226.

[15] Abd-elaziem A H, Soliman T H M. A multi-layer perceptron (mlp) neural networks for stellar classification: A review of methods and results[J]. International Journal of Advances in Applied Computational Intelligence, 2023, 3(10.54216).